\def\year{2021}\relax
\documentclass[letterpaper]{article} 
\usepackage{aaai21}  
\usepackage{times}  
\usepackage{helvet} 
\usepackage{courier}  
\usepackage[hyphens]{url}  
\usepackage{graphicx} 
\usepackage{natbib}  
\usepackage{caption} 
\usepackage{amsmath}
\usepackage{multirow}
\usepackage{booktabs}
\usepackage{amsfonts}

\title {Structure-aware Person Image Generation with\\
Pose Decomposition and Semantic Correlation}
\author {
        Jilin Tang\textsuperscript{\rm 1},
        Yi Yuan\textsuperscript{\rm 1}\thanks{Corresponding author},
        Tianjia Shao\textsuperscript{\rm 2},
        Yong Liu\textsuperscript{\rm 3},
        Mengmeng Wang\textsuperscript{\rm 3},
        Kun Zhou\textsuperscript{\rm 2}
        \\
}
\affiliations {
    \textsuperscript{\rm 1} NetEase Fuxi AI Lab \\
    \textsuperscript{\rm 2} State Key Lab of CAD\&CG, Zhejiang University \\
    \textsuperscript{\rm 3} Institute of Cyber-Systems and Control, Zhejiang University\\
    \{tangjilin, yuanyi\}@corp.netease.com, \{tjshao,mengmengwang\}@zju.edu.cn, yongliu@iipc.zju.edu.cn, kunzhou@acm.org
}

\begin{document}
	\maketitle
	
	\begin{abstract}
	    In this paper we tackle the problem of pose guided person image generation, which aims to transfer a person image from the source pose to a novel target pose while maintaining the source appearance. Given the inefficiency of standard CNNs in handling large spatial transformation, we propose a structure-aware flow based method for high-quality person image generation. Specifically, instead of learning the complex overall pose changes of human body, we decompose the human body into different semantic parts (e.g., head, torso, and legs) and apply different networks to predict the flow fields for these parts separately. Moreover, we carefully design the network modules to effectively capture the local and global semantic correlations of features within and among the human parts respectively. Extensive experimental results show that our method can generate high-quality results under large pose discrepancy and outperforms state-of-the-art methods in both qualitative and quantitative comparisons.
	\end{abstract}
	
	\section{Introduction}
\begin{figure}[t]
		\centering
		\includegraphics[width= 0.46\textwidth]{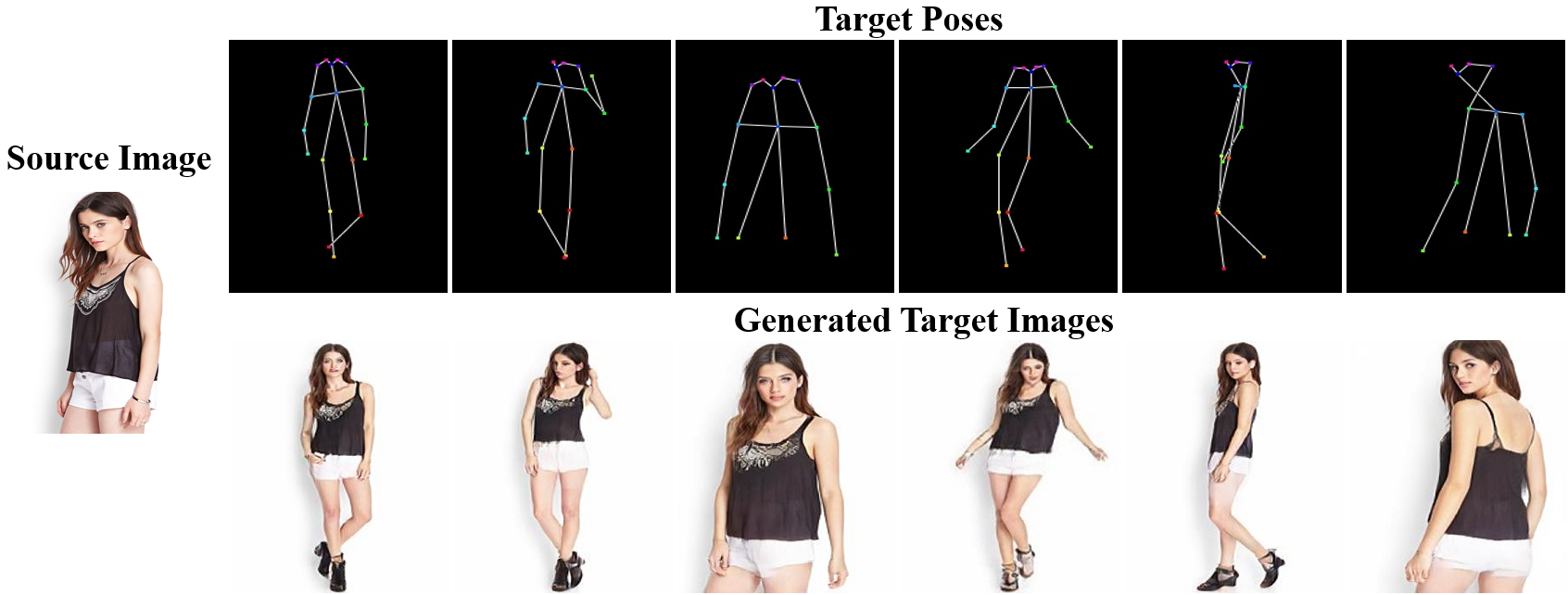}
		\caption{The generated person images
		in random target poses by our method.}
		\label{example}
	\end{figure}

	Pose guided person image generation~\cite{ma2017pose}, which aims to synthesize a realistic-looking person image in a target pose while preserving the source appearance details (as depicted in Figure~\ref{example}), has aroused extensive attention due to its wide range of practical applications for image editing, image animation, person re-identification (ReID), and so on.
	
	Motivated by the development of Generative Adversarial Networks (GANs) in the image-to-image transformation task~\cite{zhu2017unpaired}, many researchers~\cite{ma2017pose,ma2018disentangled,zhu2019progressive,men2020controllable} attempted to tackle the person image generation problem within the framework of generative models. However, as CNNs are not good at tackling large spatial transformation~\cite{ren2020deep}, these generation-based models may fail to handle the feature misalignment caused by the spatial deformation between the source and target image, leading to the appearance distortions. To deal with the feature misalignment, recently, appearance flow based methods have been proposed~\cite{ren2020deep,liu2019liquid,han2019clothflow}
	to transform the source features to align them with the target pose, modeling the dense pixel-to-pixel correspondence between the source and target features. Specifically, the appearance flow based methods aim to calculate the 2D coordinate offsets (i.e., appearance flow fields) that indicate which positions in the source features should be sampled to reconstruct the corresponding target features.
	With such flow mechanism, the existing flow based methods can synthesize target images with visually plausible appearances for most cases. However, it is still challenging to generate satisfying results when there are large pose discrepancies between the source and target images (see Figure~\ref{qua_result} for example).
	
	To tackle this challenge, we propose a structure-aware flow based method for high-quality person image generation. The key insight of our work is, incorporating the structure information can provide important priors to guide the network learning, and hence can effectively improve the results. First, we observe that the human body is composed of different parts with different motion complexities w.r.t. pose changes. Hence, instead of using a unified network to predict the overall appearance flow field of human body, we decompose the human body into different semantic parts (e.g., head, torso, and legs) and employ different networks to estimate the flow fields for these parts separately. In this way, we not only reduce the difficulty of learning the complex overall pose changes, but can more precisely capture the pose change of each part with a specific network.
	Second, for close pixels belonging to each part of human body, the appearance features are often semantically correlated. For example, the adjacent positions inside the arm should have similar appearances after being transformed to a new pose. To this end, compared to the existing methods which generate features at target positions independently with limited receptive fields, we introduce a \textit{hybrid dilated convolution block} which is composed of sequential convolutional layers with different dilation rates~\cite{yu2015multi,chen2017rethinking,li2018csrnet} to effectively capture the short-range semantic correlations of local neighbors inside human parts by enlarging the receptive field of each position. Third, the semantic correlations also exist for the features of different human parts that are far away from each other, owning to the symmetry of human body. For instance, the features of the left and right sleeves are often required to be consistent. 
	Therefore, we design a lightweight yet effective non-local component named \textit{pyramid non-local block} which combines the multi-scale pyramid pooling~\cite{he2015spatial,kim2018parallel} with the standard non-local operation~\cite{wang2018non} to capture the long-range semantic correlations across different human part regions under different scales.

	Technically, our network takes as input a source person image and a target pose, and synthesizes a new person image in the target pose while preserving the source appearance. The network architecture is composed of three modules.
	The part-based flow generation module divides the human joints into different parts, and deploys different models to predict local appearance flow fields and visibility maps of different parts respectively.
	Then, the local warping module warps the source part features extracted from the source part images, so as to align them with the target pose while capturing the short-range semantic correlations of local neighbors within the parts via the \textit{hybrid dilated convolution block}.
	Finally, the global fusion module aggregates the warped features of different parts into the global fusion features and further applies the \textit{pyramid non-local block} to learn the long-range semantic correlations among different part regions, and finally outputs a synthesized person image.
	
	The main contributions can be summarized as:
	\begin{itemize}
    	\item We propose a structure-aware flow based framework for pose guided person image generation, which can synthesize high-quality person images even with large pose discrepancies between the source and target images.
		\item 	We decompose the task of learning the overall appearance flow field into learning different local flow fields for different semantic body parts, which can ease the learning and capture the pose change of each part more precisely.
		\item 	We carefully design the modules in our network to capture the local and global semantic correlations of features within and among human parts respectively.		
	\end{itemize}
	
	\begin{figure*}[!t]
		\centering
		\includegraphics[width= 0.96\textwidth]{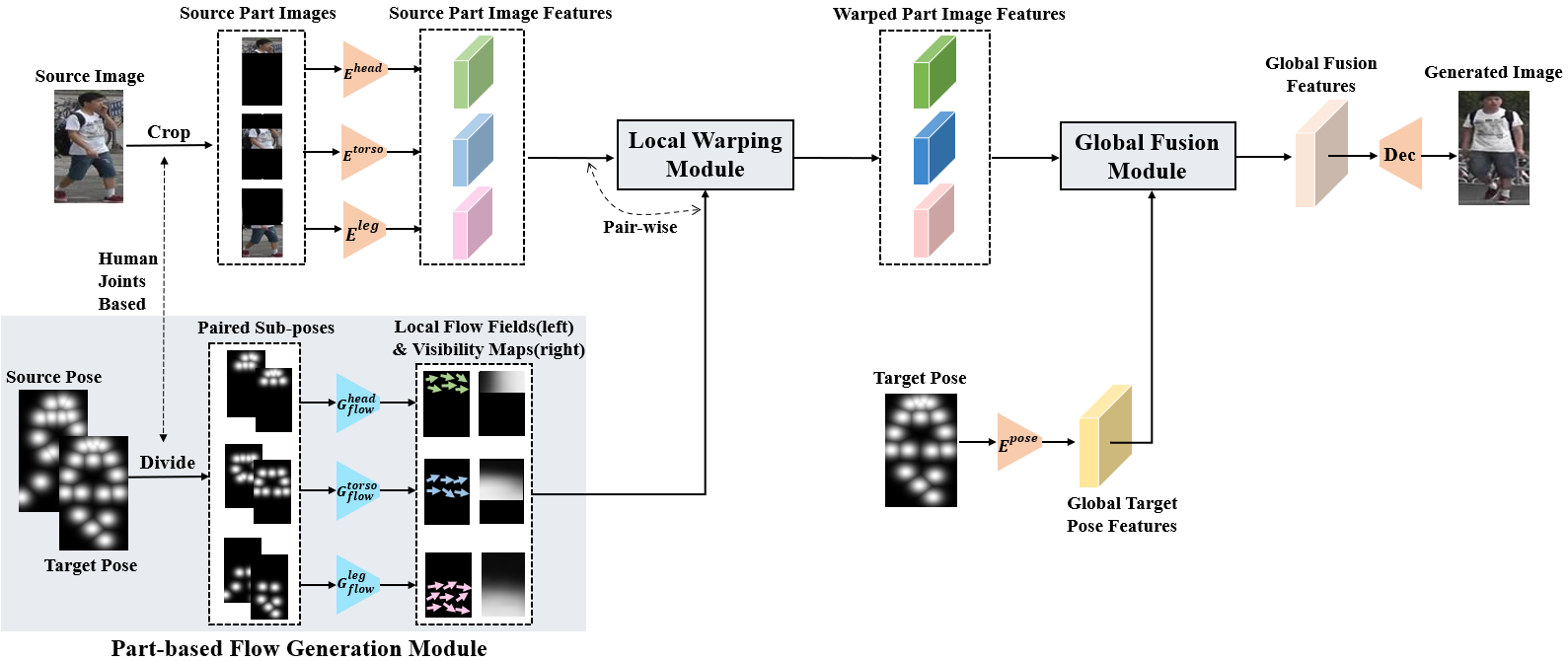}
		\caption{Overview of the proposed method. It mainly consists of three modules: the part-based flow generation module, the local warping module, and the global fusion module.}
		\label{framework}
	\end{figure*}
	
	\section{Related Work}
	Pose guided person image generation can be regarded as a typical image-to-image transformation problem~\cite{isola2017image,zhu2017unpaired} where the goal is to convert a source person image into a target person image conditioned on two constraints: (1) preserving the person appearance in the source image and (2) deforming the person pose into the target one. 
	
	Ma et al.~\cite{ma2017pose} proposed a two-stage generative network named $\rm PG^2$ to synthesize person images in a coarse-to-fine way. 
	Ma et al.~\cite{ma2018disentangled} further improved the performance of $\rm PG^2$ by disentangling the foreground, background, and pose with a multi-branch network.
	However, the both methods require a complicated staged training process and have large computation burden. 
	Zhu et al.~\cite{zhu2019progressive} proposed a progressive transfer network to deform a source image into the target image through a series of intermediate representations to avoid capturing the complex global manifold directly. However, the useful appearance information would degrade inevitably during the sequential feature transfers, which may lead to the blurry results lacking vivid appearance details. 
	Essner et al.~\cite{esser2018variational} combined the VAE~\cite{kingma2013auto} and U-Net~\cite{ronneberger2015u} to model the interaction between appearance and shape. However, the common skip connections of U-Net can't deal with the feature misalignments between the source and target pose reliably.
	To tackle this issue, Siarohin et al.~\cite{siarohin2018deformable} further proposed the deformable skip connections to transform the local textures according to the local affine transformations of certain sub-parts. However, the degrees of freedom are limited (i.e., 6 for affine), which may produce inaccurate and unnatural transformations when there are large pose changes. 
	
	Recently, a few flow-based methods have been proposed to take advantage of the appearance flow~\cite{zhou2016view,ren2019structureflow} to transform the source image to align it with the target pose. 
	Han et al.~\cite{han2019clothflow} introduced a three-stage framework named ClothFlow to model the appearance flow between source and target clothing regions in a cascaded manner. However, they warp the source image at the pixel level instead of the feature level, which needs an extra refinement network to handle the invisible contents. 
	Li et al.~\cite{li2019dense} leveraged the 3D human model to predict the appearance flow, and warped both the encoded features and the raw pixels of source image. However, they require to fit the 3D human model to all images to obtain the annotations of appearance flows before the training, which is too expensive to limit its application. 
	Ren et al.~\cite{ren2020deep} designed a global-flow local-attention framework to generate the appearance flow in an unsupervised way and transform the source image at the feature level reasonably. 
	However, this method directly takes the overall source and target pose as input to predict the appearance flow of the whole human body, which may be unable to tackle the large discrepancies between the source and target pose reliably.
	Besides, this method produces features at each target position independently and doesn't consider the semantic correlations among target features at different locations.

	\section{The Proposed Method}
	Figure~\ref{framework} illustrates the overall framework of our network. It mainly consists of three modules: the part-based flow generation module, the local warping module, and the global fusion module. 
	In the following sections, we will give a detailed description of each module.
	
	\subsection{Part-based Flow Generation Module}
	We first introduce a few notations. Let $P_{s}\in \mathbb{R}^{18\times h\times w}$ and $P_{t}\in \mathbb{R}^{18\times h\times w}$ represent the overall pose of the source image $I_{s}\in \mathbb{R}^{3\times h\times w}$ and target image $I_{t}\in \mathbb{R}^{3\times h\times w}$ respectively, where the 18 channels of the pose correspond to the heatmaps that encode the spatial locations of 18 human joints. The joints are extracted with the OpenPose~\cite{cao2017realtime}. As shown in Figure~\ref{framework}, our part-based flow generation module first decomposes the overall pose into different sub-poses via grouping 
	the human joints into different parts 
	based on the inherent connection relationship among them, 
	Then, different sub-models $G_{flow}^{local}=\left \{ G_{flow}^{head},G_{flow}^{torso},G_{flow}^{leg} \right \}$ are deployed to generate the local appearance flow fields and visibility maps of corresponding human parts respectively. Specifically, let $P_{s}^{local}=\left \{ P_{s}^{head},P_{s}^{torso},P_{s}^{leg} \right \}$ and $P_{t}^{local}=\left \{ P_{t}^{head},P_{t}^{torso},P_{t}^{leg} \right \}$ denote the decomposed source and target sub-poses, where each sub-pose corresponds to a subset of the 18 heatmaps of human joints. The sub-models $G_{flow}^{local}$ take as input $P_{s}^{local}$ and $P_{t}^{local}$, and output the local appearance flow fields $W^{local}$ and visibility maps $V^{local}$:
	\begin{equation}
	W^{local},V^{local} = G_{flow}^{local}(P_{s}^{local},P_{t}^{local}),
	\end{equation}
	where $W^{local}=\left \{ W^{head},W^{torso},W^{leg} \right \}$ records the 2D coordinate offsets between the source and target features of corresponding parts, and $V^{local}=\left \{ V^{head},V^{torso},V^{leg} \right \}$ stores confidence values between 0 and 1 representing whether the information of certain target positions exists in the source features.
	
	\subsection{Local Warping Module}
	The generated local appearance flow fields $W^{local}$ and visibility maps $V^{local}$ provide important guidance on understanding the spatial deformation of each part region between the source and target image, specifying which positions in the source features could be sampled to generate the corresponding target features. Therefore, our local warping module exploits this information to model the dense pixel-to-pixel correspondence between the source and target features. As shown in Figure~\ref{framework}, we first crop different part images from the source image, and encode them into the corresponding source part image features $F_{s}^{local}=\left \{ F_{s}^{head},F_{s}^{torso},F_{s}^{leg} \right \}$. Then, under the guidance of generated local appearance flow fields $W^{local}$, our local warping module warps $F_{s}^{local}$ to obtain the warped source features $F_{s,w}^{local}=\left \{ F_{s,w}^{head},F_{s,w}^{torso},F_{s,w}^{leg} \right \}$ aligned with the target pose. Specifically, for each target position $p=(x,y)$ in the features $F_{s,w}^{local}$, a sampling position is allocated according to the coordinate offsets $\triangle p= (\triangle x,\triangle y)$ recorded in the flow fields $W^{local}$. The features at target position are fetched from the corresponding sampling position in the source features by the bilinear interpolation. Further details are available in our supplementary material. The procedure can be written as: 
	\begin{equation}
	F_{s,w}^{local}=G_{warp}(F_{s}^{local},W^{local}).
	\end{equation}
	
	\begin{figure}[h]
		\centering
		\includegraphics[width= 0.46\textwidth]{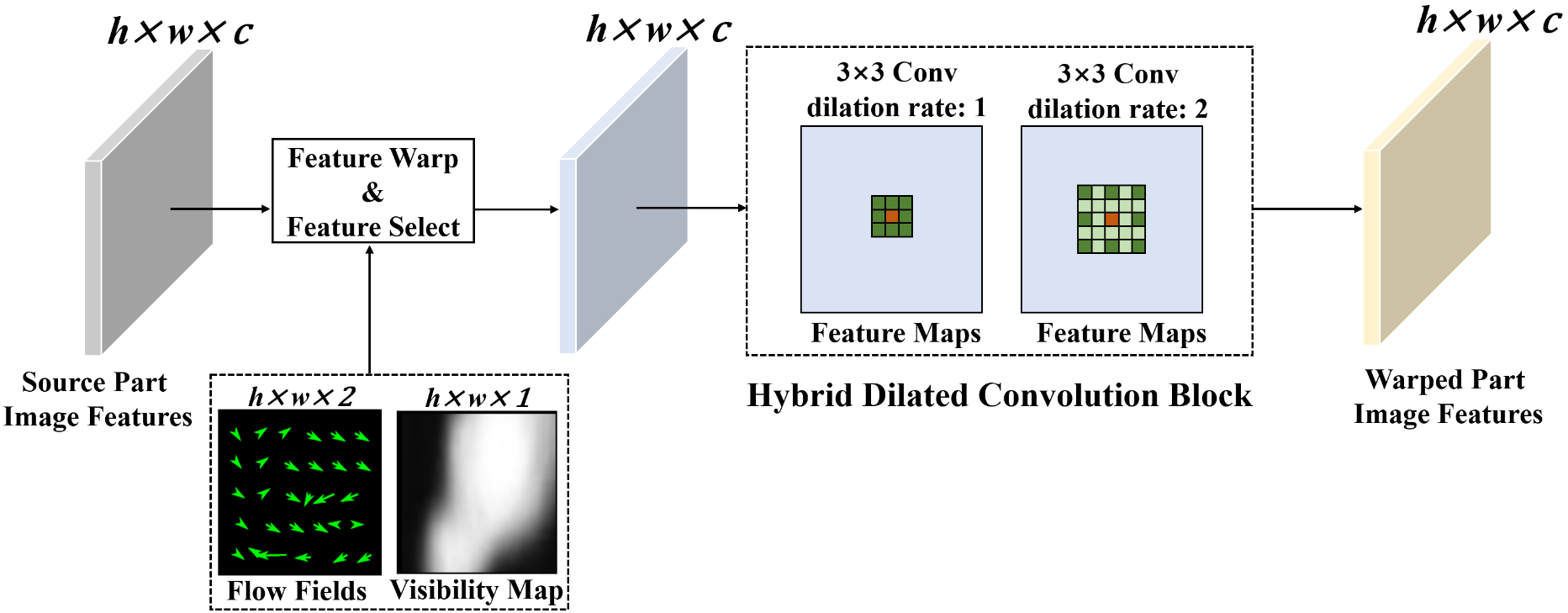}
		\caption{The local warping module. 
		 It warps the source features encoded from the corresponding part images to align them with the target pose while capturing the short-range semantic correlations of local neighbors within the parts.}
		\label{local}
	\end{figure}
	
	Considering not all appearance information of the target image can be found in the source image due to different visibilities of the source and target pose, we further take advantage of the generated local visibility maps $V^{local}$ to select the reasonable features between $F_{s,w}^{local}$ and the local target pose features $F_{pose}^{local}=\left \{ F_{pose}^{head},F_{pose}^{torso},F_{pose}^{leg} \right \}$ which are encoded from the target sub-poses. The feature selection using visibility maps is defined as:
	\begin{equation}
	F_{s,w,v}^{local} = V^{local}\cdot F_{s,w}^{local}+( 1-V^{local} )\cdot F_{pose}^{local},
	\end{equation}
	where $F_{s,w,v}^{local}=\left \{ F_{s,w,v}^{head},F_{s,w,v}^{torso},F_{s,w,v}^{leg} \right \}$ denotes the selected features for different parts.
	
	At last, in order to perceive local semantic correlations inside human parts, as shown in Figure~\ref{local}, we further introduce a \textit{hybrid dilated convolution block} which is composed of sequential convolutional layers with different dilation rates (e.g., $\left \{1, 2\right \}$ in our implementation) to capture the short-range semantic correlations of local neighbors within parts by enlarging the receptive field of each position.
	Specifically, a dilated convolution with rate $r$ can be defined as:
	\begin{equation}
	y(m,n)=\sum_{i}\sum_{j}x(m+r\times i,n+r\times j)w(i,j),
	\end{equation}
	where $y(m,n)$ is the output of dilated convolution from input $x(m,n)$, and $w(i,j)$ is the filter weight.
	Let $G_{hdcb}$ represent the \textit{hybrid dilated convolution block}. The final warped local image features of different human parts  $F_{warp}^{local}=\left \{ F_{warp}^{head},F_{warp}^{torso},F_{warp}^{leg} \right \}$ can be obtained by:
	\begin{equation}
	F_{warp}^{local} = G_{hdcb}(F_{s,w,v}^{local}).
	\end{equation}

	\subsection{Global Fusion Module}
	Let $F_{pose}^{global}$ denote the global target pose features encoded from the overall target pose $P_{t}$, which can provide additional context as to where different parts should be located in the target image. Concatenating the warped image features of different parts $F_{warp}^{local}$ and the global target pose features $F_{pose}^{global}$ together as input, the global fusion module first aggregates these local part features into the preliminary global fusion features $F_{fusion}$:
	\begin{equation}
	F_{fusion} = G_{fusion}\left (  F_{warp}^{local},F_{pose}^{global}\right ).
	\end{equation}
	
		\begin{figure}[h]
		\centering
		\includegraphics[width= 0.47\textwidth]{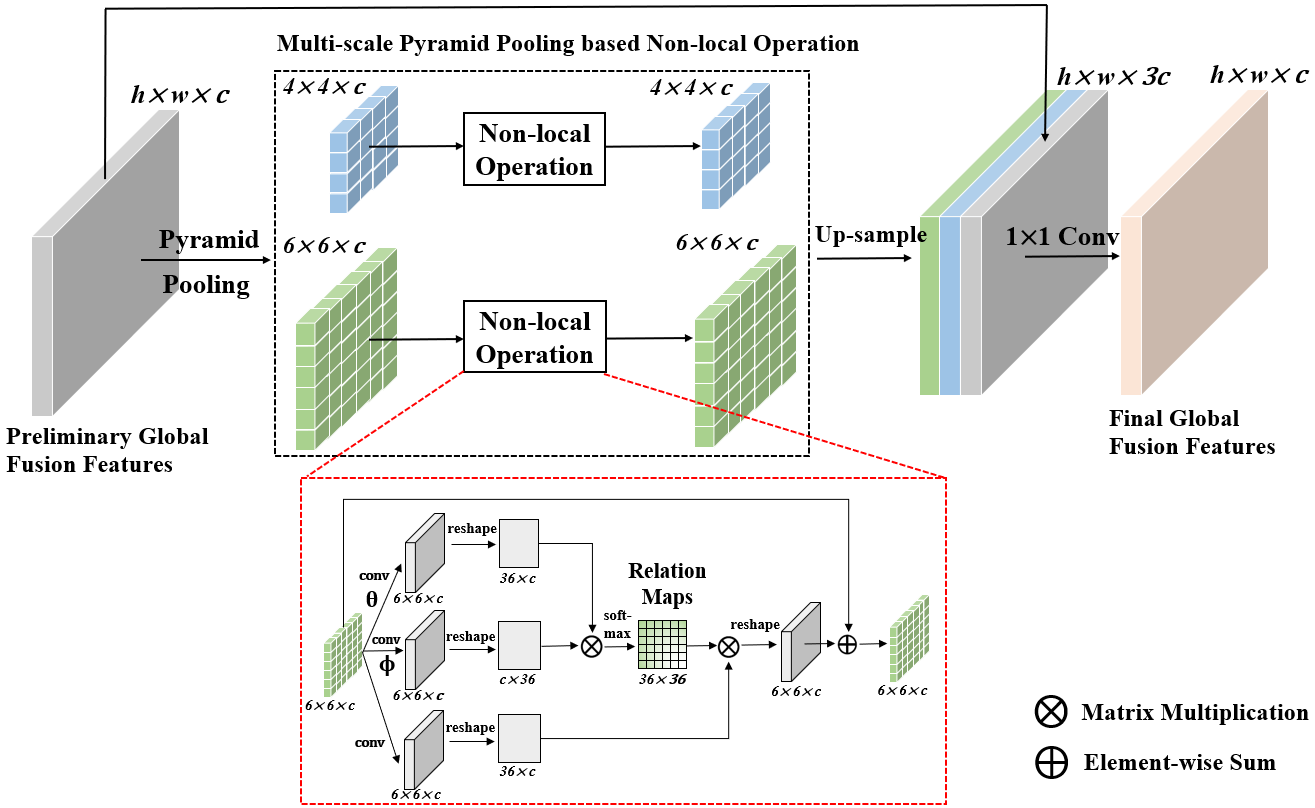}
		\caption{The global fusion module. It aggregates the warped features of different parts into the global fusion features and captures the non-local semantic correlations among different human parts.}
		\label{global}
	\end{figure}

	Due to the symmetry of human body, there can also exist important semantic correlations for the features of different human parts with long distances. 
	We therefore design a lightweight yet effective non-local component named \textit{pyramid non-local block} which incorporates the multi-scale pyramid pooling with the standard non-local operation to capture such long-range semantic correlations across different human part regions under different scales.
	Specifically, as shown in Figure~\ref{global}, given the preliminary global fusion features $F_{fusion}$, we first use the multi-scale pyramid pooling to adaptively divide them into different part regions and select the most significant global representation for each region, producing hierarchical features with different sizes (e.g., $4\times4,6\times6$) in parallel. 
	Next, we apply the standard non-local operations on the pooled features at different scales respectively to obtain the response at a target position by the weighted summation of features from all positions, where the weights are the pairwise relation values recorded in the generated relation maps (which are visualized in our experiments).
	Specifically, given the input features $x$, the relation maps $R$ are calculated by $R=softmax(\theta \left ( x \right )^{T}\phi \left ( x \right ))$, where $\theta \left ( \cdot  \right )$ and $\phi  \left ( \cdot  \right )$ are two feature embeddings implemented as $1\times1$ convolutions. 
	Let $G_{pnb}$ denote the \textit{pyramid non-local block}. The final global features $F_{global}$ are obtained via:
	\begin{equation}
	F_{global} = G_{pnb}\left ( F_{fusion} \right ).
	\end{equation}
	
	Finally, the target person image $\hat{I_{t}}$ is generated from the global features $F_{global}$ using a decoder network $Dec$ which contains a set of deconvolutional layers:
	\begin{equation}
	\hat{I_{t}}=Dec\left ( F_{global} \right ).
	\end{equation}
	
	\begin{table*}[t]
        \centering
        \resizebox{0.98\textwidth}{!}{
        \begin{tabular}{@{}c|ccccccc|cccc@{}}
        \toprule
        \multirow{2}{*}{Model} & \multicolumn{7}{c|}{Market-1501}                                                                                          & \multicolumn{4}{c}{DeepFashion}                                     \\ \cmidrule(l){2-12} 
                               & FID$\downarrow$             & LPIPS$\downarrow$           & Mask-LPIPS$\downarrow$    & SSIM$\uparrow$           & Mask-SSIM$\uparrow$    & PSNR$\uparrow$            & Mask-PSNR$\uparrow$     & FID$\downarrow$            & LPIPS$\downarrow$           & SSIM$\uparrow$           & PSNR$\uparrow$            \\ \midrule
        VU-Net                 & 24.386          & 0.3211          & 0.1747          & 0.242          & \underline{0.801}    & 13.664          & 19.102          & 13.836         & 0.2637          & \underline{0.745}    & 16.255          \\
        Def-GAN                & 29.035          & 0.2994          & 0.1496          & 0.276          & 0.793          & \underline {14.391}    & 20.425          & 26.283         & \underline {0.2330}    & \textbf{0.747} & \underline {17.524}    \\
        PATN                   & 24.917          & 0.3196          & 0.1590          & \underline {0.282}    & 0.799          & 14.241          & \underline {20.482}    & 20.399         & 0.2533          & 0.671          & 16.621          \\
        DIST                   & \textbf{21.539} & \underline {0.2817}    & \underline {0.1482}    & 0.281          & 0.796          & 14.337          & 20.421          & \textbf{7.629} & 0.2341          & 0.714          & 17.445          \\
        Ours                   & \underline {24.254}    & \textbf{0.2796} & \textbf{0.1464} & \textbf{0.290} & \textbf{0.802} & \textbf{14.526} & \textbf{20.726} & \underline {8.755}   & \textbf{0.1815} & 0.726          & \textbf{18.030} \\ \bottomrule
        \end{tabular}
        }
        \caption{Quantitative comparison with state-of-the-art methods on the Market-1501 and DeepFashion datasets. The first and second best results are bolded and underlined respectively.}
        \label{table1_2}
    \end{table*}
	
	\subsection{Training}
	We train our model in two stages. First, without the ground truth of appearance flow fields and visibility maps, we train the part-based flow generation module in an unsupervised manner using the sampling correctness loss~\cite{ren2019structureflow, ren2020deep}. Since our part-based flow generation module contains three sub-models corresponding to different parts, we train them together using the overall loss defined as:
	\begin{equation}
	L_{sam} = L_{sam}^{head} + L_{sam}^{torso} + L_{sam}^{leg},
	\end{equation}
	where $L_{sam}^{head}$,$L_{sam}^{torso}$, and $L_{sam}^{leg}$ denote the sampling correctness loss for each part respectively. The sampling correctness loss constrains the appearance flow fields to sample positions with similar semantics via measuring the similarity between the warped source features and ground truth target features. Refer to the supplementary material for details. 
	
	Then, with the pre-trained part-based flow generation module, we train our whole model in an end-to-end way. The full loss function is defined as:
	\begin{equation}
	L=\lambda _{1}L_{sam}+\lambda _{2}L_{rec}+\lambda _{3}L_{adv}+\lambda _{4}L_{per}+\lambda _{5}L_{sty},
	\end{equation}
	where $L_{rec}$ denotes the reconstruction loss which is formulated as the L1 distance between the generated target image $\hat{I_{t}}$ and ground truth target image $I_{t}$,
	\begin{equation}
	L_{rec}=\left \| I_{t} - \hat{I_{t}} \right \|_{1}.
	\end{equation}
	
	$L_{adv}$ represents the adversarial loss~\cite{goodfellow2014generative} which uses the discriminator $D$ to promote the generator $G$ to synthesize the target image with sharp details,
	\begin{equation}
	L_{adv}=\mathbb{E}\left [ log(1-D(G(I_{s},P_{s},P_{t}))) \right ]+\mathbb{E}\left [ logD(I_{t}) \right ].
	\end{equation}
	
	$L_{per}$ denotes the perceptual loss~\cite{johnson2016perceptual} formulated as the L1 distance between features extracted from special layers of a pre-trained VGG network,
	\begin{equation}
	L_{per}=\sum_{i}\left \| \phi _{i}(I_{t})-\phi _{i}(\hat{I_{t}}) \right \|_{1},
	\end{equation}
	where $\phi _{i}$ is the feature maps of the i-th layer of the VGG network pre-trained on ImageNet~\cite{russakovsky2015imagenet}.
	
	$L_{sty}$ denotes the style loss~\cite{johnson2016perceptual} which uses the Gram matrix of features to calculate the style similarity between the images,
	\begin{equation}
	L_{sty}=\sum_{j}\left \| G _{j}^{\phi}(I_{t})- G _{j}^{\phi}(\hat{I_{t}}) \right \|_{1},
	\end{equation}
	where $ G _{j}^{\phi}$ is the Gram matrix constructed from features $\phi _{j}$.
	
	\paragraph{Implementation Details.} Our model is implemented in the PyTorch framework
	using one NVIDIA GTX 1080Ti GPU with 11GB memory. We adopt the Adam optimizer ($\beta _{1}=0,\beta _{2}=0.99$)~\cite{kingma2014adam} to train our model and the learning rate is fixed to 0.001 in all experiments. For the Market-1501 dataset~\cite{zheng2015scalable}, we train our model using the images with resolution of $128\times64$, and the batch size is set to 8. 
	For the DeepFashion dataset~\cite{liu2016deepfashion}, our model is trained using the images with resolution of $256\times256$, and the batch size is 6.

    \begin{figure*}[t]
		\centering
		\includegraphics[width= 0.98\textwidth]{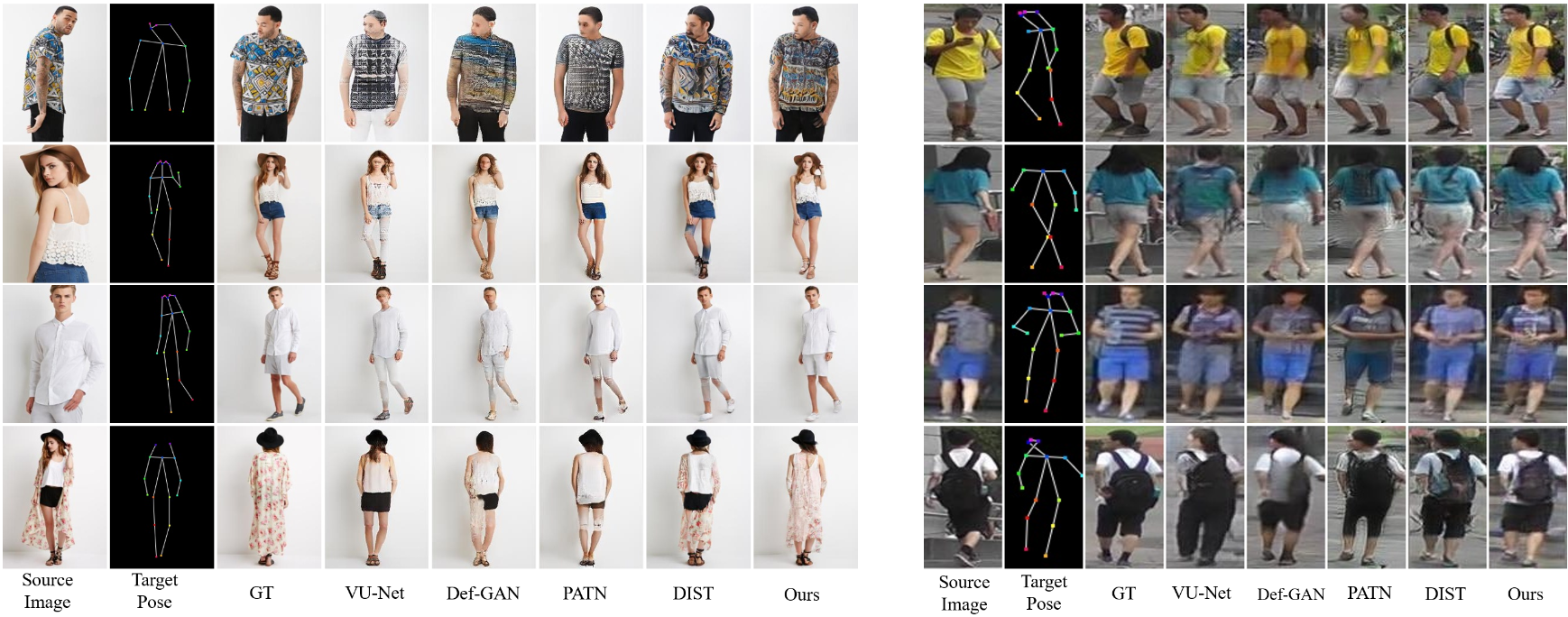}
		\caption{Qualitative comparison with state-of-the-art methods on the DeepFashion(left) and Market-1501(right) datasets.}
		\label{qua_result}
	\end{figure*}
	
	\section{Experiment}
	In this section, we perform extensive experiments to demonstrate the superiority of the proposed method over state-of-the-art methods. Furthermore, we conduct the ablation study to verify the contribution of each component in our model.
	
	\paragraph{Datasets.} We conduct our experiments on the ReID dataset Market-1501~\cite{zheng2015scalable} and the In-shop Clothes Retrieval Benchmark DeepFashion~\cite{liu2016deepfashion}. The Market-1501 dataset contains 32,668 low-resolution images ($128\times64$) which vary enormously in the pose, background, and illumination. Meanwhile, the DeepFashion dataset contains 52,712 person images ($256\times256$) with various appearances and poses. For a fair comparison, we split the two datasets following the same setting in~\cite{ren2020deep}. Consequently, we pick 263,632 training pairs and 12,000 testing pairs for the Market-1501 dataset. For the DeepFashion dataset, we randomly select 101,966 pairs for training and 8,570 pairs for testing. 
	
	\paragraph{Metrics.} It remains an open problem to evaluate the quality of generated images reasonably.
	Following the previous works~\cite{siarohin2018deformable,zhu2019progressive,ren2020deep}, we use the common metrics such as Learned Perceptual Image Patch Similarity (LPIPS)~\cite{zhang2018unreasonable}, Fr$\acute{e}$chet Inception Distance (FID)~\cite{heusel2017gans}, Structural Similarity (SSIM)~\cite{wang2004image}, and Peak Signal-to-noise Ratio (PSNR) to assess the quality of generated images quantitatively. Specifically, both LPIPS and FID calculate the perceptual distance between the generated images and ground truth images in the feature space w.r.t. each pair of samples and global distribution, respectively. Meanwhile, SSIM and PSNR indicate the similarity between paired images in raw pixel space. For the Market-1501 dataset, we further calculate the masked results of these metrics to exclude the interference of the backgrounds. Furthermore, considering that these quantitative metrics may not fully reflect the image quality~\cite{ma2017pose}, we perform a user study to qualitatively evaluate the quality of generated images.

	\subsection{4.1	Comparison with State-of-the-art Methods}
	\paragraph{Quantitative Comparison.}
	As shown in Table~\ref{table1_2}, we compare our method with four state-of-the-art methods including VU-Net~\cite{esser2018variational}, Def-GAN~\cite{siarohin2018deformable}, PATN~\cite{zhu2019progressive}, and DIST~\cite{ren2020deep} on the Market-1501 and DeepFashion datasets. Specifically, we download the pre-trained models of state-of-the-art methods and evaluate their performance on the testing set directly. As we can see, our method outperforms the state-of-the-art methods in most metrics on both datasets, demonstrating the superiority of our model in generating high-quality person images.

	\paragraph{Qualitative Comparison.}
	Figure~\ref{qua_result} shows the qualitative comparison of different methods on the two datasets. All the results of state-of-the-art methods are obtained by directly running their pre-trained models released by authors. 
	As we can see, for the challenging cases with large pose discrepancies (e.g., the first two rows on the left of Figure~\ref{qua_result}), the existing methods may produce results with heavy artifacts and appearance distortion.
	In contrast, for the DeepFashion dataset~\cite{liu2016deepfashion}, our model can generate realistic images in arbitrary target poses, which not only reconstructs the reasonable and consistent global appearances, but preserves the vivid local details such as the textures of clothes and hat. Especially, our model is able to produce more suitable appearance contents for target regions which are invisible in the source image such as the legs and backs of clothes (see the last three rows). 
	For the Market-1501 dataset~\cite{zheng2015scalable}, our model yields natural-looking images with sharp appearance details whereas the artifacts and blurs can be observed in the results of other state-of-the-art methods. More results can be found in the supplementary material.

	\paragraph{User Study.}
	We perform a user study to judge the realness and preference of the images generated by different methods. For the realness, we recruit 30 participants to judge whether a given image is real or fake within a second. Following the setting of previous work~\cite{ma2017pose,siarohin2018deformable,zhu2019progressive}, for each method, 55 real images and 55 generated images are selected and shuffled randomly. Specifically, the first 10 images are used to warm up and the remaining 100 images are used to evaluate. For the preference, in each group of comparison, a source image, a target pose, and 5 result images generated by different methods are displayed to the participants, and the participants are asked to pick the most reasonable one w.r.t. both the source appearance and target pose. We enlist 30 participants to take part in the evaluation and each participant is asked to finish 30 groups of comparisons for each dataset. 
	As shown in Table~\ref{table3}, our method outperforms the state-of-the-art methods in all subjective measurements on the two datasets, especially for the DeepFashion dataset~\cite{liu2016deepfashion} with higher resolution, verifying that the images generated by our model are more realistic and faithful.
	
	\begin{table}[h]
		\centering
		\resizebox{0.31\textwidth}{!}{
		\begin{tabular}{@{}c|cccc@{}}
			\toprule
			\multirow{2}{*}{Model} & \multicolumn{2}{c|}{Market-1501}                 & \multicolumn{2}{c}{DeepFashion} \\ \cmidrule(l){2-5} 
			& G2R$\uparrow$            & \multicolumn{1}{c|}{Prefer$\uparrow$} & G2R$\uparrow$            & Prefer$\uparrow$     \\ \midrule
			VU-Net                 & -\,-             & 11.44                           & -\,-             & 1.00           \\
			Def-GAN                & 41.03          & 10.00                           & 5.23  & 1.44           \\
			PATN                   & 38.03          & 14.00                           & 10.93          & 2.22           \\
			DIST                   & 47.37          & 23.11                           & 38.30          & 28.89          \\
			Ours                   & \textbf{50.00} & \textbf{41.45}                  & \textbf{43.83} & \textbf{66.45} \\ \bottomrule
		\end{tabular}
		}
	\caption{User study($\%$). \textbf{G2R} means the percentage of generated images rated as real w.r.t. all generated images. \textbf{Prefer} denotes the user preference for the most realistic result among different methods.}
	\label{table3}
	\end{table}

	\subsection{Ablation Study}
	We further perform the ablation study to analyze the contribution of each technical component in our method. We first introduce the variants implemented by alternatively removing a corresponding component from our full model.
	
	\noindent \textbf{w/o the part-based decomposition (w/o Part).} This model removes the part-based decomposition in our flow generation module, and directly estimates the whole flow field of human body to warp the global source image features.
	
	\noindent \textbf{w/o the hybrid dilated convolution block (w/o HDCB).} This model removes the \textit{hybrid dilated convolution block} in our local warping module, and directly uses the selected part features to conduct the subsequent feature fusion.
	
	\noindent \textbf{w/o the pyramid non-local block (w/o PNB).} This model removes the \textit{pyramid non-local block} in our global fusion module, and simply takes the preliminary global fusion features as input to generate the final target images.
	
	\noindent \textbf{Full.} This represents our full model. 
	
	Table~\ref{table4} shows the quantitative results of ablation study on the DeepFashion dataset~\cite{liu2016deepfashion}. We can see that, our full model achieves the best performance on all evaluation metrics except SSIM, and the removal of any components will degrade the performance of the model.
	
	\begin{table}[h]
		\centering
		\resizebox{0.32\textwidth}{!}{
		\begin{tabular}{@{}c|cccc@{}}
			\toprule
			\multirow{2}{*}{Model} & \multicolumn{4}{c}{DeepFashion}                                     \\ \cmidrule(l){2-5} 
			& FID$\downarrow$            & LPIPS$\downarrow$           & SSIM$\uparrow$           & PSNR$\uparrow$            \\ \midrule
			w/o Part                & 13.736         & 0.2090          & 0.716         & 17.420          \\
			w/o PNB                & 9.302          & 0.1832          & 0.728          & 17.945          \\
			w/o HDCB               & 9.326          & 0.1829          & \textbf{0.729} & 18.021          \\
			Full                   & \textbf{8.755} & \textbf{0.1815} & 0.726          & \textbf{18.030} \\ \bottomrule
		\end{tabular}
		}
		\caption{The quantitative results of ablation study on the DeepFashion dataset. The best results are bolded.}
		\label{table4}
	\end{table}

\begin{figure}[h]
		\centering
		\includegraphics[width= 0.45\textwidth]{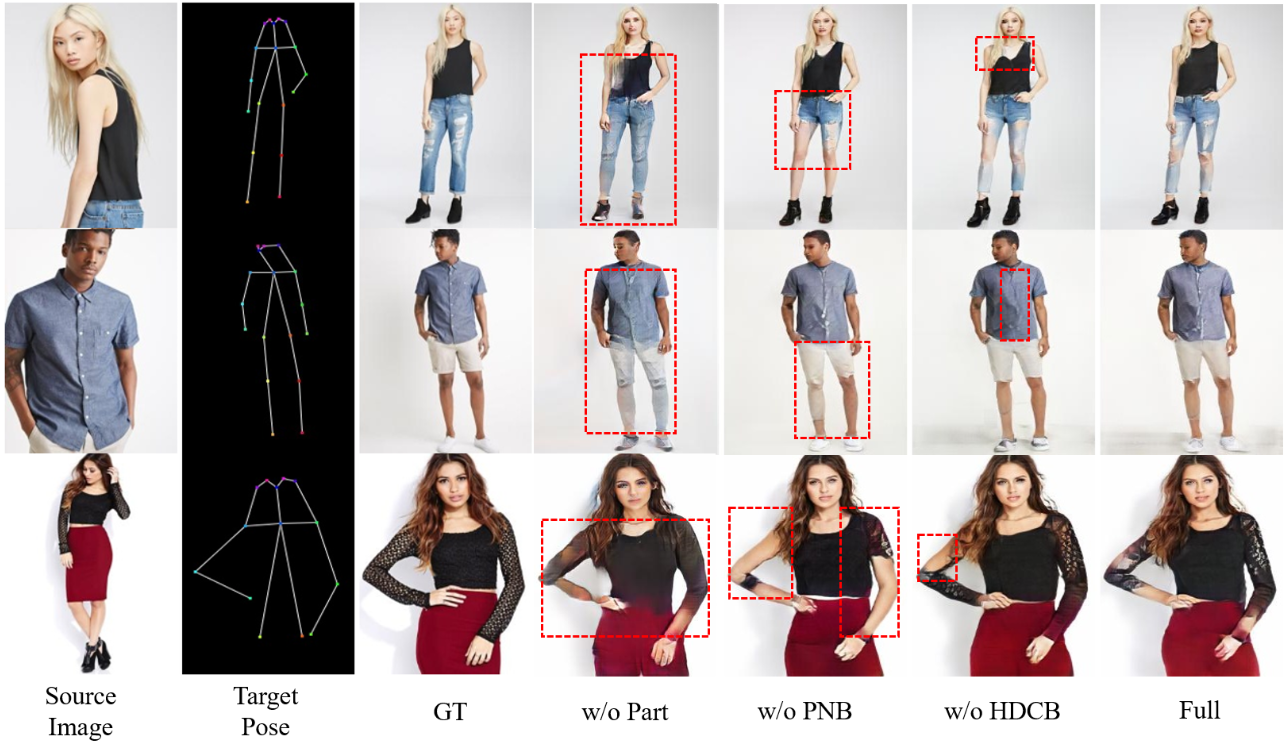}
		\caption{The qualitative comparison of ablation study.}
		\label{aba}
	\end{figure}

	Qualitative comparison of different ablation models is demonstrated in Figure~\ref{aba}. 
	We can see that, although the models w/o Part, w/o PNB, and w/o HDCB can generate target images with correct poses, they can't preserve the human appearances in source images very well.
	Specifically, there exists heavy appearance distortion on the results produced by the model w/o Part, because of the difficulty in directly learning the overall flow fields of human body under large pose discrepancies.
	The results generated by the model w/o PNB often suffer from the inconsistency in global human appearance since it doesn't explicitly consider the long-range semantic correlations across different human parts. Besides, the images produced by the model w/o HDCB may lose some local appearance details because it can't fully capture the short-range semantic correlations of local neighbors within a certain part. In contrast, our full model can reconstruct the most realistic images which not only possess consistent global appearance, but maintain vivid local details.
	
	\subsection{Visualization of The Relation Map}
\begin{figure}[!h]
		\centering
		\includegraphics[width= 0.45\textwidth]{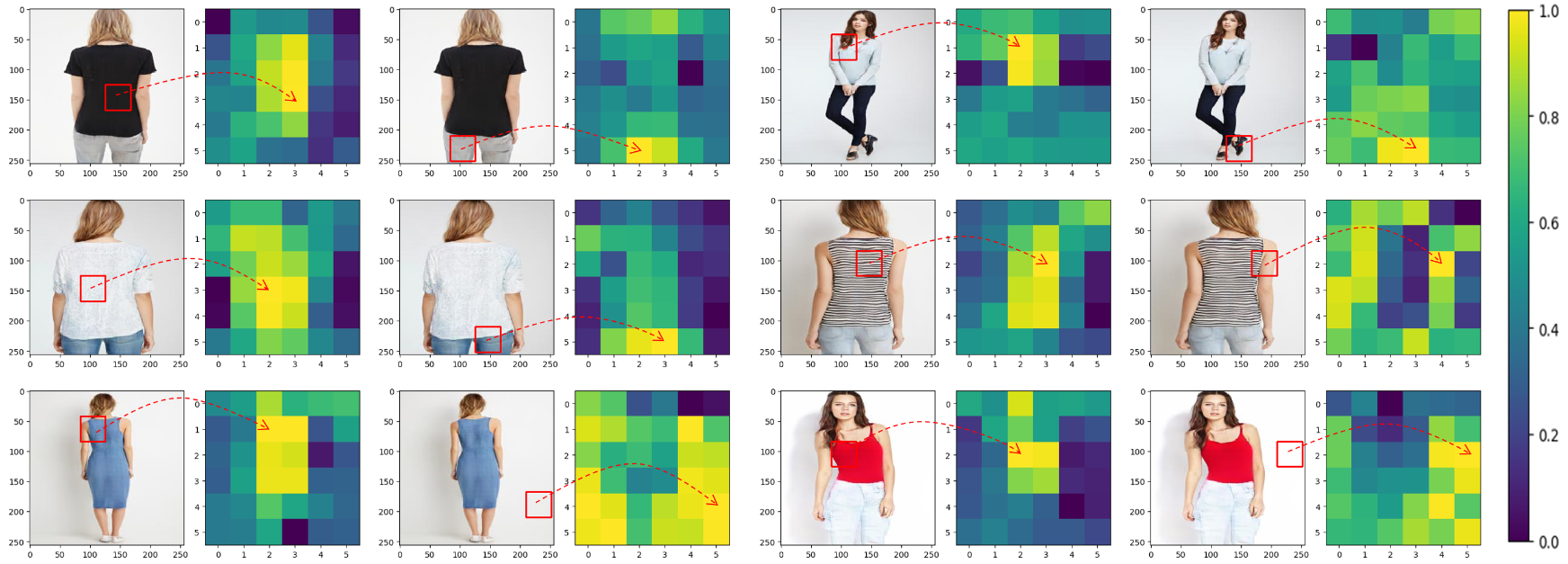}
		\caption{Visualization of the relation map w.r.t. a certain target patch marked by a red rectangle in the image.}
		\label{vis}
	\end{figure}

	To illustrate the effectiveness of our \textit{pyramid non-local block} in capturing the global semantic correlations among different human parts, in Figure~\ref{vis} we visualize the generated relation map (e.g., size of $6\times6$), which represents the relation values of all patches w.r.t a certain target patch. 
	As we can see, for a target patch in a certain image region (e.g., shirt, pants, background), the patches with similar semantics usually have larger relation values w.r.t. this target patch, indicating that our \textit{pyramid non-local block} can capture the non-local semantic correlations among different part regions effectively.
	
	\subsection{Person Image Generation in Random Poses}
\begin{figure}[h]
		\centering
		\includegraphics[width=0.46\textwidth]{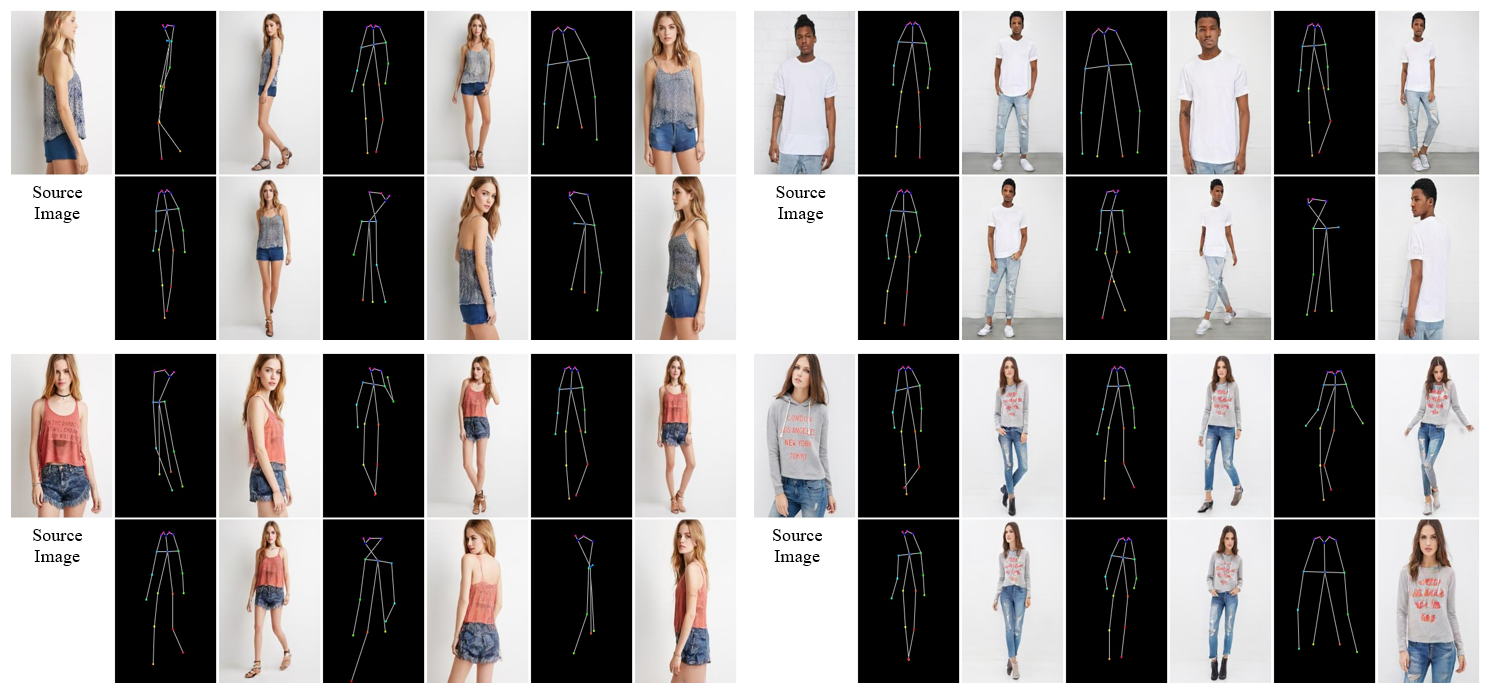}
		\caption{The results of generated person images in random target poses on the DeepFashion dataset.}
		\label{random_fashion}
	\end{figure}

	As shown in Figure~\ref{random_fashion}, given the same source person image and a set of target poses selected from the testing set randomly, our model is able to generate the target images with both vivid appearances and correct poses 
	, demonstrating the versatility of our model sufficiently.

	\section{Conclusion}
	We present a structure-aware appearance flow based approach to generate realistic person images conditioned on the source appearances and target poses. 
	We decompose the task of learning the overall appearance flow field into learning different local flow fields for different human body parts, which can simplify the learning and model the pose change of each part more precisely.
	Besides, we carefully design different modules within our framework to capture the local and global semantic correlations of features inside and across human parts respectively. Both qualitative and quantitative results demonstrate the superiority of our proposed method over the state-of-the-art methods. Moreover, the results of ablation study and visualization verify the effectiveness of our designed modules.
    
    \section{Acknowledgments}
    This work is supported by the National Key R\&D Program of China (2018YFB1004300).
    
	\bibliography{ref}
	
\end{document}